\pdfoutput=1
\documentclass[11pt]{article}

\usepackage[T1]{fontenc}

\usepackage[utf8]{inputenc}

\newcommand\cryc{Cryptonite}

\usepackage[hyperref]{emnlp2021}
\usepackage{times}
\usepackage{latexsym}
\usepackage[capitalize,noabbrev]{cleveref}

\usepackage{graphicx}
\usepackage{xcolor}
\usepackage{multirow}
\usepackage{makecell}
\usepackage{booktabs} 
\usepackage{subcaption}
\usepackage{siunitx}
\usepackage{xurl}  % for auto-breaking url lines
\usepackage{enumitem}  % for setting space between list items
\usepackage{makecell}  % for making multirow cells
\usepackage{arydshln}

\usepackage{microtype}

\title{Cryptonite: A Cryptic Crossword Benchmark\\for Extreme Ambiguity in Language}

\author{Avia Efrat\thanks{\hspace{0.6em}Equal contribution.} \hspace{1.6em} Uri Shaham\footnotemark[1] \hspace{1.6em} Dan Kilman \hspace{1.4em} Omer Levy \\
        Tel Aviv University \\
        \texttt{\{avia.efrat,uri.shaham1\}@gmail.com}}

\date{}

\begin{document}
\maketitle

\begin{abstract}
Current NLP datasets targeting ambiguity can be solved by a native speaker with relative ease.
We present \cryc{}, a large-scale dataset based on \textit{cryptic} crosswords, which is both linguistically complex and naturally sourced.
Each example in Cryptonite is a \textit{cryptic clue}, a short phrase or sentence with a misleading surface reading, whose solving requires disambiguating semantic, syntactic, and phonetic wordplays, as well as world knowledge.
Cryptic clues pose a challenge even for experienced solvers, though top-tier experts can solve them with almost 100\% accuracy.
\cryc{} is a challenging task for current models; fine-tuning T5-Large on 470k cryptic clues achieves only 7.6\% accuracy, on par with the accuracy of a rule-based clue solver (8.6\%).
\end{abstract}

\section{Introduction} \label{sec:introduction}

The ambiguity of natural language is one of the most fundamental challenges in NLP research.
While there are works and datasets specifically targeting ambiguity \citep{Levesque2011TheWS,raganato-etal-2017-word,Sakaguchi2020WINOGRANDEAA}, these can be solved by a native speaker with relative ease.
Can we design a dataset with ambiguities that pose a challenge even to competent native speakers?

\begin{figure}[t]
  \centering
  \includegraphics[width=0.9\columnwidth]{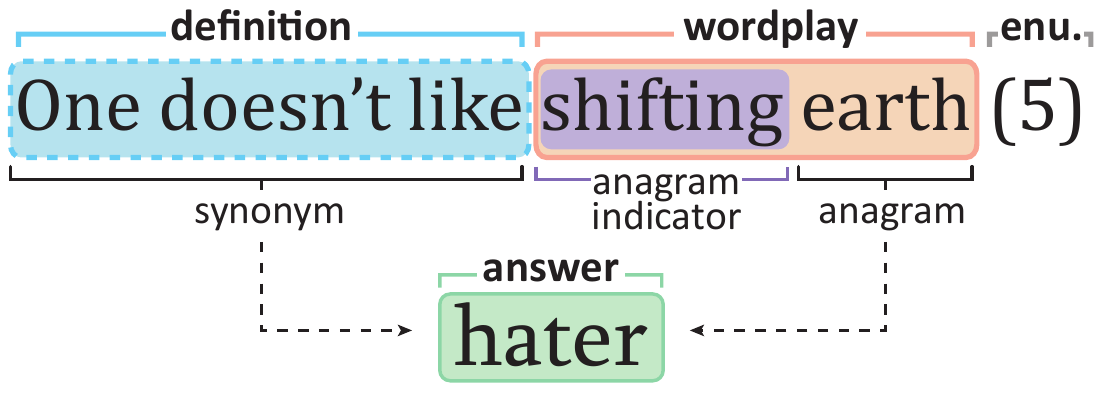}
  \label{fig:ExampleClueAnagram}
  \caption{\textbf{How to solve the cryptic clue ``One doesn't like shifting earth (5)'':} Solving usually starts by figuring out which of the clue's words belong to the \textit{definition} (blue) and which to the \textit{wordplay} (orange).
  Next, one needs to figure out the \textit{type} of wordplay, which is often hinted by an \textit{indicator} (purple).
  In our case, ``shifting'' hints that the answer is an anagram of some part of the wordplay.
  As the \textit{enumeration} (gray) states the answer is a five-letter word, ``earth'' is a promising candidate for anagraming.
  Finally, given that ``hater'' is both an anagram of ``earth'' and a synonym of the definition, we conclude it to be the correct answer.}
  \vspace{-10pt}
\end{figure}

We present \cryc{}, a large-scale dataset based on \textit{cryptic} crosswords, which is both linguistically complex and naturally sourced.
\cryc{}'s 523K examples are taken from professionally-authored cryptic crosswords, making them less prone to artifacts and biases than examples created by crowdsourcing \citep{gururangan-etal-2018-annotation,geva-etal-2019-modeling}.
Each example in \cryc{} is a \textit{cryptic clue}, a short phrase or sentence with a misleading surface reading, which poses a challenge even for humans experienced in cryptic crossword solving.
A cryptic clue usually consists of two underlying parts: \textit{wordplay} and \textit{definition}.
The clue's answer is both a disambiguation of the wordplay and, at the same time, directly answers the definition.
While solving cryptic clues requires disambiguating semantic, syntactic, and phonetic wordplays, as well as world knowledge, clues are designed to have only one possible answer.
See \cref{fig:ExampleClueAnagram} for an example clue and its solution. 

We provide a standard baseline by fine-tuning the generic T5-Large conditional language model \citep{2020t5} on \cryc{}, achieving only 7.6\% accuracy.
For comparison, a rule-based cryptic clue solver \citep{robin_deits_2021_4541744} achieves 8.6\% accuracy.
These results highlight the challenge posed by \cryc{}, making it a candidate for assessing the disambiguation capabilities of future models.

Analyzing the results of both baselines, we find a correlation between performance on individual clues and a human assessment of the clue's difficulty,
and that the enumeration (answer length) is highly informative.
Finally, we show that ensuring that the answers of train and test examples are mutually exclusive is critical for a candid estimation of T5's ability to solve cryptic clues in general.

\section{Cryptic Crosswords} \label{sec:CrypticCrosswords}

A cryptic crossword, just like a regular (non-cryptic) crossword, is a puzzle designed to entertain humans, comprised of \textit{clues} whose answers are to be filled in a letter grid.
Unlike regular crosswords, where answers are typically synonyms or hyponyms of the clues \citep{severyn-etal-2015-distributional}, \textit{cryptic} clues have a misleading surface reading, and solving them requires disambiguating wordplays.
A cryptic clue has only one possible answer, even when taken outside the context of the letter grid.\footnote{For example, a regular crossword can have a clue like ``A Sesame Street character (4)''.
Without existing letters from the grid, a solver cannot determine if the answer is ``elmo'' or ``bert''.
This cannot happen with a cryptic clue, which by design does not require any letter grid information for solving.}

Generally, a cryptic clue (henceforth, ``clue'') consists of two parts: \textit{wordplay} and \textit{definition}.
The wordplay-definition split is not given to the solver, and parsing it is usually the first step in solving a clue.
Both the wordplay and the definition lead to the answer, but each in a different manner.
While the definition is directly related to the answer (e.g. a synonym or a hypernym), the wordplay needs to be deciphered, usually with the help of an \textit{indicator} that hints at the wordplay's type.

\begin{figure}[htbp] 
  \begin{subfigure}{\linewidth}
    \centering
    \includegraphics[width=0.9\columnwidth]{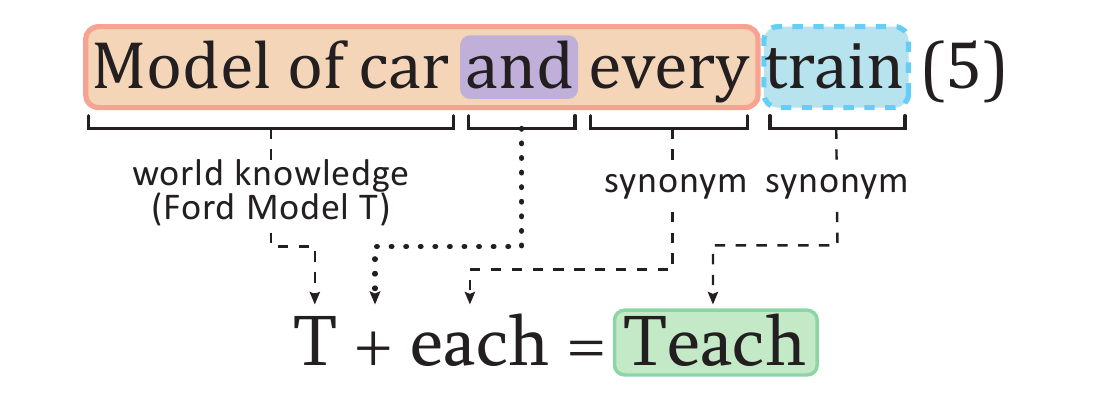} 
    \caption{A clue with a relatively simple \textit{additive} wordplay, also requiring world knowledge.
    One can decipher the wordplay by identifying ``and'' as a concatenation indicator.} 
    \label{fig:ClueExamples:a} 
    \vspace{4ex}
  \end{subfigure}
  \begin{subfigure}{\linewidth}
    \centering
    \includegraphics[width=0.9\columnwidth]{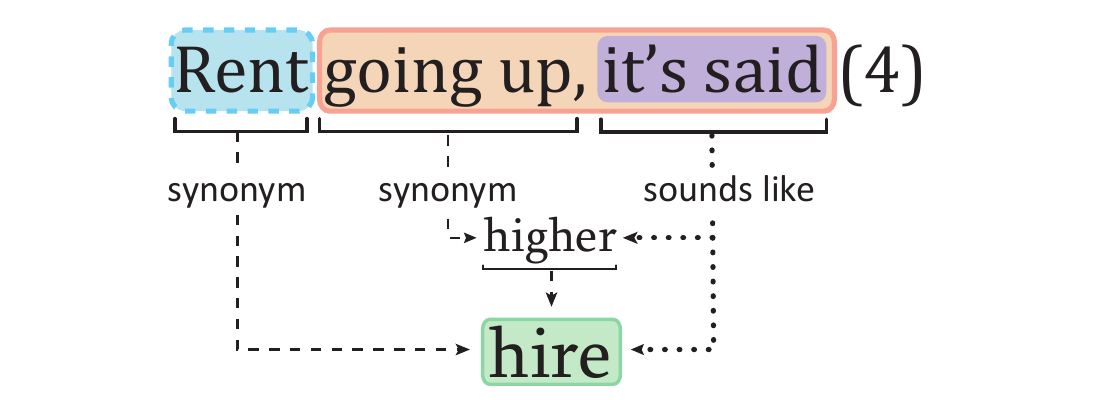} 
    \caption{Clues can also have \textit{phonetic} wordplays.
    Here, ``it's said'' implies that the wordplay is a homophone of the answer.}
    \label{fig:ClueExamples:b} 
    \vspace{4ex}
  \end{subfigure}
  \begin{subfigure}{\linewidth}
    \centering
    \includegraphics[width=0.9\columnwidth]{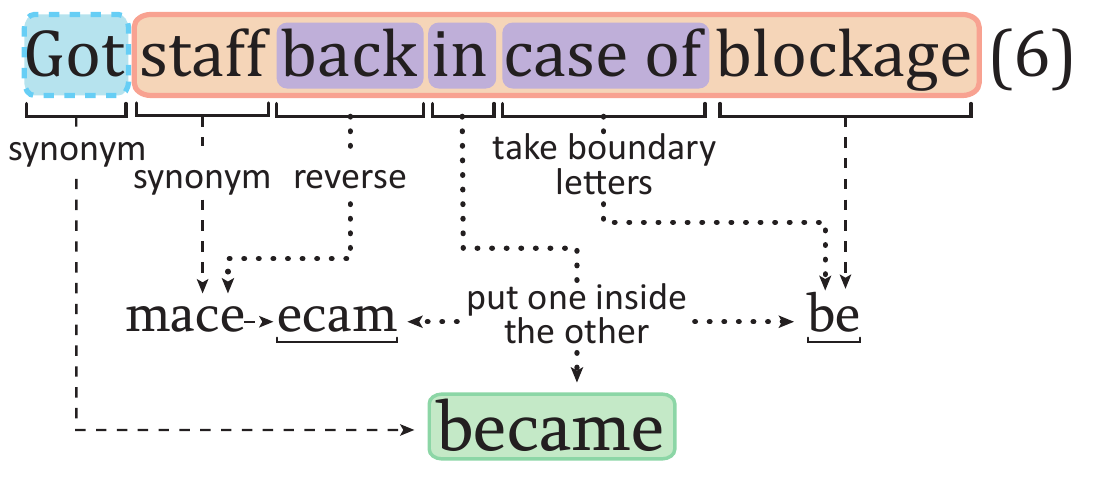} 
    \caption{Many clues combine more than one type of wordplay.
    This clue composes three: \textit{reversing} the letters of a word (``back''), and \textit{inserting} it (``in'') into the \textit{boundary letters} of another (``case of'').}
    \vspace{4ex}
    \label{fig:ClueExamples:c} 
  \end{subfigure}
  \begin{subfigure}{\linewidth}
    \centering
    \includegraphics[width=0.9\columnwidth]{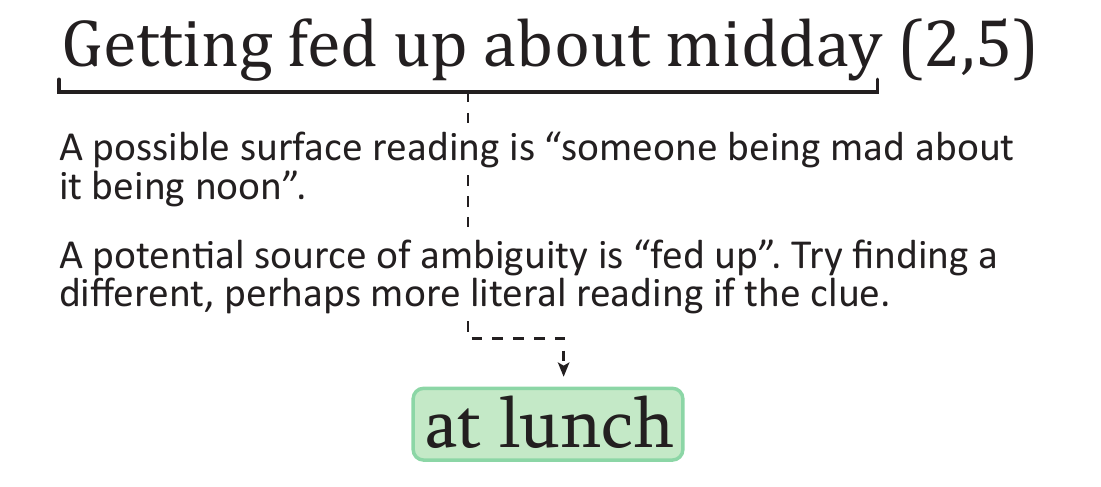} 
    \caption{Although many wordplays can be roughly clustered into types and deciphered based on indicators, there is no silver bullet for solving cryptic crosswords.
    For this clue, even the standard wordplay-definition split does not apply;
    instead, the entire clue points to the answer.}
    \vspace{2ex}
    \label{fig:ClueExamples:d}
  \end{subfigure} 
  \caption{Various examples of cryptic clues and how to solve them. Answers are in green, definitions in light blue (dashed frame), wordplays in orange, and indicators are in purple.}
  \label{fig:ClueExamples} 
  
\end{figure}

\cref{fig:ClueExamples} walks through several clues with different types of wordplays and their solution.
These examples provide only a glimpse into the rich and diverse world of cryptic crosswords. For a deeper dive, see the guidebook by \citet{moorey2018times}.

\section{Dataset}
\label{sec:dataset}

We introduce \cryc{}, a dataset of 523,114 cryptic clues from 17,375 English-language crosswords published in The Times\footnote{\url{https://www.thetimes.co.uk/puzzleclub/crosswordclub/home/crossword-cryptic}} and The Telegraph\footnote{\url{https://puzzles.telegraph.co.uk/crossword-puzzles/cryptic-crossword}} between October 2000 and October 2020.\footnote{We do not redistribute the dataset, but provide code for creating it with subscriptions to The Times and The Telegraph:  \url{https://github.com/aviaefrat/cryptonite}}

For preprocessing, we remove clue-answer duplicates and examples whose answer and enumeration do not match.
In addition, we remove any examples with the same clue but with a different answer. 
While this occurred in less than 0.1\% of the data, these examples violate the principle that a clue must have a single solution once the wordplay is deciphered (see \cref{sec:CrypticCrosswords}).

We follow the recent findings of \citet{lewis2020question}, and split \cryc{} into train, validation, and test sets, where no answer is shared between them.
Answer splitting creates a far more challenging benchmark for supervised models than naive random splits (see \cref{subsec:Analysis}).
\cref{tab:DatasetStatistics} shows some basic statistics of the final \cryc{} dataset.%\footnote{To respect intellectual property, we provide a script for downloading the data. Please do not use the dataset for commercial purposes and do not distribute it without permission.}

\begin{table}[t]
\small
\centering
\begin{tabular}{@{}lcccc@{}}
    \toprule
         &  \multirow{2}{*}{\textbf{Examples}} & \textbf{Unique} & \textbf{Clue} &  \textbf{Answer} \\
       & & \textbf{Answers} & \textbf{Length} & \textbf{Length} \\
    \midrule
    \textit{train} & 470,803 & 80,837 & 7.76 & 1.22 \\
    \textit{valid} & ~~26,156 & ~~4,534 & 7.77 & 1.20 \\
    \textit{test} & ~~26,157 & ~~4,538  & 7.77 & 1.24 \\
    \bottomrule
\end{tabular}
\caption{Overview of \cryc{}. Reported lengths are mean values. Words are delimited by spaces.}
\label{tab:DatasetStatistics}
\end{table}

\section{Experiments}

We provide initial results on \cryc{} using two baselines: T5-Large \cite{2020t5} and a rule-based cryptic clue solver \cite{robin_deits_2021_4541744}.
Despite training on half a million clues (T5) or being tailored to the task (rule-based solver), both approaches solve only a small portion of the test data, demonstrating that \cryc{} is indeed a challenging task.
We further investigate two properties of the data: how difficulty (as perceived by humans) correlates with accuracy, and the informativeness of enumeration.
In addition, we analyze how a naive data split affects the performance of T5, demonstrating that partitioning by answers is crucial for obtaining a candid estimate of the neural model's ability to generalize to new cryptic clues.

\subsection{Baselines}

\paragraph{T5-Large}
Following current NLP methodology, we fine-tune the 770M parameter T5-Large \cite{2020t5} on \cryc{}.
The model's encoder takes the clue as input, and uses the decoder to predict the answer using teacher forcing during training and beam search ($b=5$) during inference.

We use HuggingFace \citep{wolf-etal-2020-transformers} with the recommended settings \citep{2020t5}, optimizing with AdaFactor \citep{pmlr-v80-shazeer18a} at a constant learning rate of 0.001.
We train until convergence with a patience of 10 epochs and a batch size of 7000 tokens, selecting the best model checkpoint using validation set accuracy.

T5 uses SentencePiece tokenization \citep{kudo-richardson-2018-sentencepiece}, which might incur some information loss, as many clues require character-level manipulations.

\paragraph{Rule-based Solver}
We also gauge the abilities of a rule-based solver with a manually-crafted probabilistic grammar \cite{robin_deits_2021_4541744}.
Building on the assumption that a clue can usually be split into a wordplay and a definition (\cref{sec:CrypticCrosswords}), the solver tries to find the most probable parse such that the wordplay yields a semantically-similar result to the definition.
The similarity between the definition and the parsed wordplay is calculated using expert-authored resources such as WordNet \citep{10.1145/219717.219748}.
Some less frequent wordplay types, such as \textit{homophones} (\cref{fig:ClueExamples:b}) and \textit{hidden-at-intervals} \cite[Chapter 3]{moorey2018times}, are not implemented in the solver's grammar.

\subsection{Main Benchmark}

\begin{table}[t]
  \small
  \centering
  \begin{tabular}{@{}lcc@{}}
    \toprule
 \multirow{2}{*}{\textbf{Baseline}} &  \multicolumn{2}{c}{\textbf{Accuracy}}  \\
   &  \textbf{Validation}  & \textbf{Test} \\
    \midrule
    T5-Large & 7.44\% & 7.64\% \\
    Rule-based Solver & 8.26\% & 8.58\% \\
    \bottomrule
  \end{tabular}
  \caption{Baseline performance on \cryc{}. Accuracy is measured using exact string match.} 
  \label{tab:BaselineResults}
\end{table}

We first evaluate our baselines on the main dataset.
\cref{tab:BaselineResults} shows that both approaches are able to solve a small portion of the clues.
Even though the T5 model is trained on roughly half a million examples, it does not exceed the performance of the rule-based solver.
For comparison, top-tier human experts are able to solve even very hard clues with almost 100\% accuracy \citep{10.3389/fpsyg.2016.00567, 10.3389/fpsyg.2018.00904},
though this expertise is acquired through significant training.
\cref{appx:ExamplePredictions} shows a selection of examples and the respective predictions of T5.

\subsection{Analysis}
\label{subsec:Analysis}

\paragraph{Correlation with human perception of difficulty}
\textit{Quick} cryptic crosswords is a subgenre of cryptic crosswords aimed at beginners, with clues designed to be easier to solve.
\cryc{}'s test set contains 2,081 such clues.
Examining the results of our main benchmark, \cref{tab:QuickVsNonQuick} shows that both baselines perform better on quick clues, suggesting a correlation between human assessment of linguistic difficulty and the models' performance on clues.\footnote{All quick clues are taken from \href{https://www.thetimes.co.uk/article/times-crosswords-introducing-the-new-quick-cryptic-gj6fk59cfkh}{Times Quick Cryptic} (TQC). For a fair comparison to the quick clues, we consider a clue as non-quick only if it was published in The Times after March 10th 2014 (when TQC was introduced), and not as a part of TQC. \cryc{}'s test set contains 4,653 such non-quick clues.}

\begin{table}[t]
\small
\centering
\begin{tabular}{@{}lcc@{}}
  \toprule
    & \textbf{Quick Clues} & \textbf{Non-Quick Clues} \\
  \midrule
  T5-Large          & 12.83\%  &  3.40\% \\
  Rule-based Solver & 13.50\%  &  5.78\% \\
  \bottomrule
\end{tabular}
\caption{Accuracy on quick cryptic clues vs non-quick cryptic clues. Both baselines perform better on clues that were deemed easier by human experts.}
\label{tab:QuickVsNonQuick}
\end{table}

\paragraph{The effect of enumeration}
The enumeration is the number (or numbers) in parentheses at the end of a clue indicating the number of letters in its answer, e.g. (7) or (5,4).
To measure the informativeness of enumeration, we run our main experiment again, this time without providing the enumeration.
\cref{tab:WithEnumerationVsWithoutEnumeration} shows an accuracy drop in both baselines when the enumeration is not provided.\footnote{\cryc{}'s metadata contains additional information that could help a solver, such as \textit{orientation} (whether the clue is across or down in the grid).
Knowing the orientation can help in finding the clue's wordplay-definition split.}
While it is to be expected that enumeration helps the rule-based solver, we see that T5 is able to leverage this information as well. 

\begin{table}[t]
\small
\centering
\begin{tabular}{@{}lcc@{}}
  \toprule
    & \textbf{With Enum.} & \textbf{Without Enum.} \\
  \midrule
  T5-Large          & 7.64\%  &  4.90\% \\
  Rule-based Solver & 8.58\%  &  3.58\% \\
  \bottomrule
\end{tabular}
\caption{Comparison of baseline accuracy when enumeration is provided and when it is not provided.}  
\label{tab:WithEnumerationVsWithoutEnumeration}
\end{table}

\begin{figure}[t]
\includegraphics[width=\columnwidth]{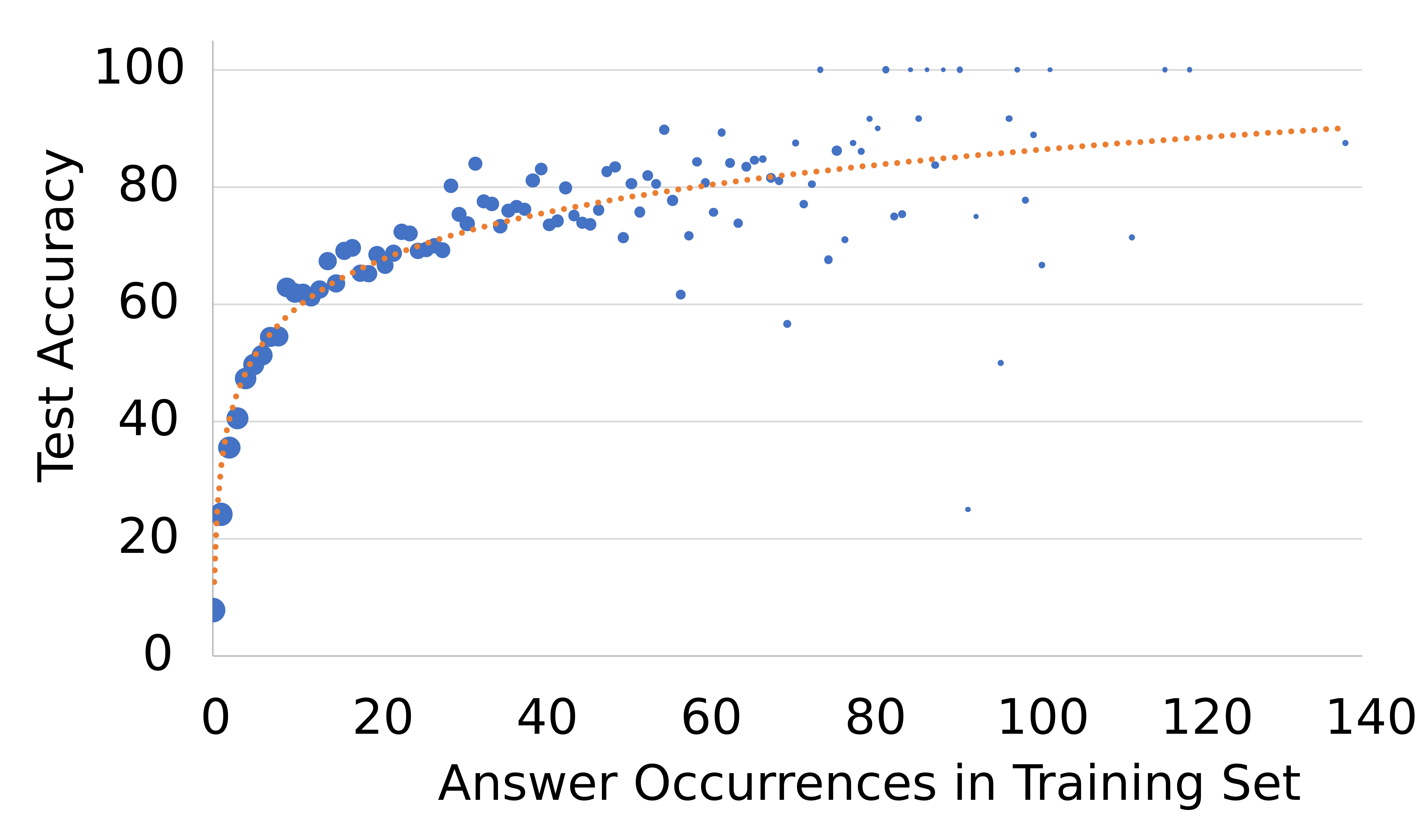}
\caption{The naive split exhibits a strong correlation between T5's accuracy on clues from the test set (vertical axis) and the number of times their answer appears in the train set (horizontal).
Each dot's size represents the number of clues from the test set whose answer appears in the train set $n$ times.
Trend line is logarithmic.}\label{fig:answer-in-train}
\end{figure}

\paragraph{Why do we split the data by answer?}
Many clues that share the same answer are paraphrases of each other (\cref{appx:SimilarClues}).
A neural model such as T5 might exploit this information and by copying answers from memorized training examples.
Therefore, to test whether a model has learnt a general process for solving cryptic clues, we follow \citet{lewis2020question} and make \cryc{}'s default split the \textit{answer split}, in which the answers of the train, validation, and test sets are mutually exclusive.

We compare the answer split with a \textit{naive} (random) partition of the data.
\cref{tab:AnswerSplitVsRandomSplit} shows that a naive split of \cryc{} will grossly overestimate the performance of T5; while the rule-based solver's performance barely changes, T5-Large is now able to solve an additional 50\% of the entire test set.
Further analyzing the naive test set (\cref{fig:answer-in-train}), we observe that the probability of T5 solving a clue is highly correlated with the number of times its answer appeared in the training set.
This result indicates that that a significant part of the performance difference is due to the paraphrasing artifact,
and that ensuring unseen test answers is critical for establishing a true estimate of a model's ability to solve cryptic clues.

\begin{table}[t]
\small
\centering
\begin{tabular}{@{}lcc@{}}
  \toprule
  \textbf{Data Split}    & \textbf{T5-Large}   &
  \textbf{Rule-based Solver} \\
  \midrule
  Answer & ~~7.64\%  & 8.58\% \\
  Naive & 56.16\%   & 8.43\% \\
  \bottomrule
\end{tabular}
\caption{Comparing test set accuracy of our \textit{answer} split (where the answers of test examples do not appear during training) and the \textit{naive} random split.}
\label{tab:AnswerSplitVsRandomSplit}
\end{table}

\section{Related Work} \label{sec:RelatedWork}

\paragraph{Cryptic crosswords}
\citet{10.1093/comjnl/22.1.67} attempt to devise a formal language for describing cryptic clues.
\citet{HART199216} define four stages of rule-based solving, and implement the second stage -- ``syntactic identification''.
In our work we focus on creating a large-scale dataset of a cryptic clues and apply neural and rule-based methods to establish a strong baseline.
\citet{Hardcastle2001Using,hardcastle2007riddle} focuses on rule-based approaches for \textit{creating} cryptic clues given a word as an answer.
Although in our work we test solving abilities, the reverse direction of creating a clue from an answer is also challenging, and the \cryc{} dataset could prove useful in this direction as well.

\paragraph{Language disambiguation}
In addition to works and datasets specifically targeting disambiguation on the word level \citep{Levesque2011TheWS,raganato-etal-2017-word,Sakaguchi2020WINOGRANDEAA}, there are other domains strongly related to language disambiguation.
Among them are pun disambiguation \citep{miller-gurevych-2015-automatic,miller-etal-2017-semeval}, and sarcasm detection \citep{10.1145/3124420,oprea-magdy-2020-isarcasm}.
However, to the best of our knowledge \cryc{} is the first dataset both large in scale (unlike pun disambiguation), and containing a variety of wordplays (unlike sarcasm detection).

\paragraph{Non-cryptic crosswords}
As described in \cref{sec:CrypticCrosswords}, non-cryptic (``regular'') crosswords are the common crosswords found in most newspapers. There are works introducing regular crossword datasets, some even containing a small percentage of more ``tricky'' clues\footnote{Compare \texttt{Florida fruit (6) $\rightarrow$ orange} to \texttt{Where to get a date (4) $\rightarrow$ palm.}} \citep{littman2002probabilistic}. However, identifying this small portion of clues requires human effort, whereas \cryc{} is already guaranteed to consist entirely of cryptic clues. In addition, works on solving regular crosswords typically rely on an external database of clues \citep{ernandes2005webcrow,barlacchi-etal-2014-learning,severyn-etal-2015-distributional}. When given a clue as an input, these systems search the database for the most similar clues, in hope they share the answer with the input clue. In \cryc{}, the answers of the train, validation, and test sets are mutually exclusive (\cref{sec:dataset}). In doing so, we hope to shift the focus of solving from memorization to reasoning, which is especially interesting in the setting of \textit{cryptic} clues.

\section{Conclusion}

We presente \cryc{}, a large-scale dataset based on cryptic crosswords, whose solving requires disambiguating a variety of wordplays.
We saw that the standard approach of fine-tuning T5-Large on \cryc{} does not outperform an existing rule-based model, achieving 7.6\% and 8.6\% accuracy respectively, while human experts achieve close to 100\% accuracy.
These results highlight the challenge posed by \cryc{}, and will hopefully encourage further research on disambiguation tasks that are not easily solved by a native speaker.
\section*{Acknowledgements}

This work was supported by the Tel Aviv University Data Science Center, the Blavatnik Fund, the Alon Scholarship, and Intel Corporation.
We are grateful for the help of Avia Amitay in graphical design, Lior Ben-Moshe in infrastructure, and for the detailed feedback of Mor Geva.

\bibliographystyle{acl_natbib}
\bibliography{anthology,custom}

\appendix

\newpage

\section{Example Predictions}
\label{appx:ExamplePredictions}

\begin{table}[htbp]
\small
\centering
\begin{tabular}{@{}ll@{}}
    \toprule
    \textbf{Clue} &
    \textbf{Answer} \\
    \midrule
    Act like tragic heroine with cold extremity & mimic \\
    Group of musicians prohibited on the radio & band \\
    Assumed diamonds to be shelved & put on ice \\
    Is in control of distant armies abroad & administrates \\
    Second parasite & tick \\
    \bottomrule
\end{tabular}
\caption{Examples of correct predictions of T5-Large.}
\label{tab:ExamplesOfCorrectPredictions}
\end{table}

\begin{table}[htbp]
\small
\centering
\begin{tabular}{@{}lll@{}}
    \toprule
    \textbf{Clue} &
    \textbf{Answer} &
    \textbf{Prediction} \\
    \midrule
    Suffer death at riverside & endure & strand \\
    Travel free heading for eastbourne & ride & trip \\
    Toast gordon! & brown & gobbi \\
    Pair of braces & four  & pair \\
    \multirow{2}{*}{Performing insect begged for food} & eggs & beef \\
    & benedict & wellington \\
    \bottomrule
\end{tabular}
\caption{Examples incorrect predictions of T5-Large.}
\label{tab:ExamplesOfWrongPredictions}
\end{table}
\section{Similar Clues}
\label{appx:SimilarClues}

\begin{table}[htbp]
\small
\centering
\begin{tabular}{@{}lc@{}}
    \toprule
    \textbf{Clue} &
    \textbf{Answer} \\
    \midrule
    greek upper chamber &
    \multirow{2}{*}{attic} \\
    greek for 'upper room' & \\
    \cmidrule{1-2}
    flat race maybe failing to finish & \multirow{2}{*}{even} \\
    flat race possibly unfinished &  \\
    \cmidrule{1-2}
    one playing minor part in run & \multirow{2}{*}{extra} \\
    one with small part in film run? & \\
    \cmidrule{1-2}
    think to make changes in partner &
    \multirow{2}{*}{meditate} \\
    contemplate change in partner & \\
    \cmidrule{1-2}
    beginning assault &
    \multirow{2}{*}{onset} \\
    start of an attack & \\
    \cmidrule{1-2}
    what we learn by accepting established award &
    \multirow{2}{*}{rosette} \\
    what we learn by accepting fixed award & \\
    \cmidrule{1-2}
    animal in forest, a grizzly &
    \multirow{2}{*}{stag} \\
    animal in forest, a gazelle & \\
    \bottomrule
\end{tabular}
\caption{Examples of pairs of similar clues from the naive split, one from the train set and one from the test set.}
\label{tab:ClueParaphraseExamples}
\end{table}

\end{document}